\documentclass{article}

\usepackage{PRIMEarxiv}

\usepackage[utf8]{inputenc} 
\usepackage[T1]{fontenc}    
\usepackage{hyperref}       
\usepackage{url}            
\usepackage{booktabs}       
\usepackage{amsfonts}       
\usepackage{nicefrac}       
\usepackage{microtype}      
\usepackage{lipsum}
\usepackage{fancyhdr}       
\usepackage{graphicx}       
\graphicspath{{media/}}     
\usepackage{color,array}
\usepackage{graphicx}
\usepackage{amsmath}
\usepackage{multirow}
\usepackage{subcaption}
\usepackage{caption}
\usepackage{booktabs}
\usepackage[table]{xcolor}
\usepackage{balance}

\title{Enhancing Abnormality identification: Robust Out-Of-Distribution strategies for Deepfake Detection}

\author{
  Luca Maiano, Fabrizio Casadei, and Irene Amerini \\
  Department of Computer, Control and Management Engineering (DIAG) \\
  Sapienza University of Rome \\
  Rome, Italy \\
  \texttt{\{maiano,amerini\}diag@uniroma1.it, casadei.1952529@studenti.uniroma1.it}\\
}

\begin{document}
\maketitle

\begin{abstract}
Detecting deepfakes has become a critical challenge in Computer Vision and Artificial Intelligence. Despite significant progress in detection techniques, generalizing them to open-set scenarios continues to be a persistent difficulty. Neural networks are often trained on the closed-world assumption, but with new generative models constantly evolving, it is inevitable to encounter data generated by models that are not part of the training distribution. To address these challenges, in this paper, we propose two novel Out-Of-Distribution (OOD) detection approaches. The first approach is trained to reconstruct the input image, while the second incorporates an attention mechanism for detecting OODs. Our experiments validate the effectiveness of the proposed approaches compared to existing state-of-the-art techniques. Our method achieves promising results in deepfake detection and ranks among the top-performing configurations on the benchmark, demonstrating their potential for robust, adaptable solutions in dynamic, real-world applications.
\end{abstract}


\section{Introduction}
\label{section_introduction}
Traditional machine learning systems are typically trained under the closed-world assumption, an assumption that assumes that a machine learning system primarily encounters data similar to those it was trained on, often known as In-Distribution (ID) data. However, this assumption is limiting. In the real world, data can be dynamic and unpredictable, introducing regions into the model that have not yet been explored. Deepfakes are a prime example of this phenomenon. Generative AI techniques are advancing very quickly, making it virtually impossible to train a model on any generator, as generators are introduced over time. In such a scenario, it makes much more sense to adopt the open-world assumption, which aims to identify test samples that belong to a different distribution than the training one, known as Out-Of-Distribution (OOD). The presence of OOD data can cause a significant drop in system performance in real-world applications. In the context of deepfakes, OOD detection can indicate when retraining of the model is needed to handle previously unseen samples that would otherwise be misinterpreted~\cite{TASSONE2024104143}. Integrating OOD detection systems into the AI-driven media identification process offers significant benefits, improving the adaptability, robustness, and overall effectiveness of detection systems.

Detecting deepfakes is one of the most complex challenges in the field of digital media forensics~\cite{amerini2021deep}, which is concerned with analyzing and verifying the authenticity of digital content. This challenge is amplified by two main factors: (i) the rapid progress of generation techniques and (ii) the heterogeneous nature of manipulations. On the one hand, technological advances lead to the development of increasingly sophisticated models capable of synthesizing extremely realistic visual content, making it difficult to distinguish deepfakes from authentic media. On the other hand, the variety of generative techniques used to create fake content introduces additional complexities. These techniques include face swapping, altering facial expressions (face reprocessing), manipulating facial attributes (face retouching), full-face synthesis, and even generating non-face-related content. This diversity of approaches requires the development of versatile and robust detection methods that can cope with the constant evolution of manipulation technologies. For this reason, there are several techniques to perform Deepfake detection~\cite{tariang2024synthetic}, which vary based on the type of data processed, the overall architecture of the model in action and the nature of the features exploited, such as color~\cite{df_color}, GAN fingerprint~\cite{df_gan_fingerprint}, PRNU~\cite{df_prnu}, self-attention mechanism~\cite{df_SA} or based on neural behaviors~\cite{df_neuron_behaviors}. While some works use image-based algorithms (unimodal), as in our case, much research has been conducted on techniques based on video frames and those that include both audio and video data (multimodal)~\cite{dfd_multimodal_1, dfd_multimodal_2, dfd_multimodal_3}. While video-based approaches are often more powerful than image-based ones, their applicability is limited to specific types of content.


In this work, we treat deepfake detection as a problem of recognizing data from an unknown distribution. This results in two cooperating models for deepfake and OOD detection tasks. Both models address a binary classification problem, whose inputs can be categorized as \emph{Real} or \emph{Fake} and as \emph{In distribution} or \emph{Out of distribution}. Both tasks share intrinsic challenges related to designing a solution that can generalize over known and unknown distributions, lack of training data considering the complexity and dynamic shifting of data domains, and the need for low computational complexity to solve these problems.

Following the guidelines in \cite{ood_survey} which characterizes five sub-tasks all under the generalized OOD framework, our task is represented at best under Anomaly Detection, given problem specifics like binary detection and pure covariate shift. 
The proposed contributions of this work are different. First, the designing of and-hoc pre-processing on a existing dataset, providing the possibility to customize settings and build different deepfake detection dataset instances. This data partitioning is particularly convenient to perform both deepfake and Out-Of-Distribution detection, in this way, the dataset can be used as a new hard OOD benchmark. Second, we propose a pipeline composed of an In-Distribution module and an Out-Of-Distribution detector, which collaborates and shares auxiliary data. The In-Distribution module's aim is to identify deepfake samples at the image level, while OOD detector discriminates between images that belong or not to the training distribution. The OOD detection solution respects the reconstruction-based and output-based paradigms. Two distinct approaches are presented, involving Convolutional models and Vision Transformers, and three slightly different OOD detector architectures are described and integrated with both approaches. Models are trained and tested on the custom dataset. Finally, our models are trained and tested also on a popular OOD benchmark, to give a wider look at the complete panorama of known OOD detection techniques, and understand better the potentialities of this methodology, analyzing the pros and cons.\\

\section{State of the art}
\label{section_state_of_the_art}
\noindent
\subsection{Deepfake Detection}
Nowadays, the advent of deepfake technology has introduced unprecedented challenges in the protection of both privacy and security. Deepfake detection in visual media is a task that can be addressed by following different approaches and exploiting different types of clues and intrinsic data information. The overall approach to this task can follow both unimodal and multimodal strategies. An ensemble approach \cite{df_ensemble} is a useful tool for combining many unimodal models, such as those for audio or visual data.
Various image-based approaches perform the identification of GAN-originated media taking into account convolution traces. Huang et al. \cite{df_gan_image} used a gray-scale fakeness map to take advantage of the artifacts in the upsampling process in GAN-generated deepfakes. To further increase the robustness of the model, the attention mechanism, partial data augmentation, and individual sample clustering are used by L. Guarnera et al. \cite{df_conv_pattern}, who present a technique for extracting a collection of local features, meant to mimic the convolutional patterns commonly seen in images using an expectation maximization approach. Authors \cite{df_background} specifically note that the deepfake generator adjusts the face area while preserving the context. Thus, it is feasible to easily identify fake media by concentrating on potential differences between faces and backgrounds.
The usage of a Convolutional Vision Transformer (CViT) to detect Deepfakes is introduced by the authors in \cite{df_cvit}.
Abady et al. \cite{siamese_based_dfd} exploit the Siamese Network architecture to verify whether two synthetic input images are produced by the same generative model. In addition, they propose a methodology to verify a possible claim about the technique used to generate a synthetic image.
In the article published by Tariang et al. \cite{synt_image_detection} are discussed the most effective techniques for synthetic image detection and attribution. Additionally, they compare closed-set and open-set scenarios for attribution at the architecture level, evaluating the performance of various methods.
Authors in \cite{OOD_abstrain} exploit BNNs combined with Noise Contrastive Prior Estimation demonstrating the effectiveness of this approach in a Deepfake Detection task with OOD data. Differently from our approach, they presented three abstain techniques useful in the presence of unknown data to avoid false decisions, and make the ID model robust to OOD data. The model training is characterized in a one-stage setting to both learn the multi-class task and to force high uncertainty in the OOD region.\\
One of the common problems in the detection of forgeries is the overfitting of the model on the training data, and the consequent lack of generalization across different datasets and deepfake techniques. Korshunov et al. \cite{df_fewshot} investigate techniques to improve the generalization of deepfake detection methods and propose a methodology that includes data augmentation and few-shot tuning methods. The work in \cite{df_gaussian} proposes DDT (Deep Distribution Transfer) a novel zero- and few-shot transfer learning method for facial forgery detection that models data with a multimodal distribution.

\noindent
\subsection{Out-Of-Distribution Detection}
Another open problem in the literature, is the improvement of reliability and safety in machine learning, to avoid falling victim of overconfident results over data never seen. In other words, we want models that are able to generalize as much as possible. For this purpose, the well-known challenge of finding a robust deepfake detection solution, can be integrated with the research on the Out-Of-Distribution detection problem. The first work that we associate with OOD detection in the literature, is the MSP (Maximum Softmax Probability) baseline \cite{ood_baseline}. The approach involves taking the maximum softmax probability to discriminate between correctly classified examples and misclassified and out-of-distribution examples.
ODIN (Out-of-DIstribution detector for Neural Networks) \cite{ood_odin} exploits gradient information to perform OOD detection.
The confidence estimating branch method \cite{ood_conf_branch} incentivizes the neural network to produce confidence estimations that correctly reflect the model’s ability to estimate a correct prediction for any input.
After training, we can perform a threshold-based classification only based on the confidence estimation.
In a multi-label context, the techniques proposed by Want et al. \cite{ood_multi_label_energy} called JointEnergy for out-of-distribution (OOD) detection, aggregating label-wise energy scores across all labels to establish a new OOD score.
A similar approach can be applied also to a multi-class task, as proposed by Liu et al. \cite{ood_multi_class_energy}. It’s also possible to exploit visual attention heatmap \cite{ood_vit} from the Vision Transformer (ViT) classifier to perform OOD detection. Authors use the attention heatmap reconstruction error as discriminative information for the classification of OOD samples and ID examples.
We also mention the possibility of including the CutMix \cite{ood_cutmix} augmentation technique to improve training generalization and OOD detection.
Among all the techniques that don’t require additional OOD data, authors in \cite{ood_gram} show the possibility of using Gram matrices to detect OOD samples.
Output-based methods in OOD detection focus on utilizing the model’s output. This category can be further characterized by Post-hoc detection and training-based methods. Post-hoc methods don’t change the training procedure and objective \cite{ood_baseline, ood_odin, ood_multi_label_energy}.
Instead, training-based methods modify the training of the classifier at different levels like model structure, loss function, augmentation, normalization, etc. One example is G-ODIN \cite{ood_godin}, which extends ODIN \cite{ood_odin} using a specific training loss and the selection of hyperparameters on ID data, including perturbation magnitude.
Distance-based methods for Out-of-Distribution detection operate on the principle that OOD samples should be significantly distant from the centroids or prototypes of In-Distribution classes. For instance, the detection thanks to minimum Mahalanobis distance \cite{ood_mahalanobis}.
Density-based techniques for OOD identification utilize some probabilistic models to predict the In-distribution and classify test data in low-density areas as OOD. An example of this category is the mixed Gaussian distribution \cite{ood_mixed_gaussian} technique.
Bayesian models are employed in this kind of problem too, they represent uncertainty through inference based on Bayes’ rule. Bayesian neural network \cite{ood_bayesian_nn} represents the most known method of this category, based on Markov chain Monte Carlo \cite{ood_monte_carlo} to draw samples from the posterior distribution of the model.
Methods with real outlier exposure involve using a set of collected Out-Of-Distribution input, or "outliers," during training to help models learn the discrepancy between in-distribution and OOD data. MixOE \cite{ood_mixOE} proposes to mix ID and OOD images to obtain informative outliers for better regularization. Whether OOD data are not available which is very common in practice, it’s possible to use outlier data generation (i.e. via GANs).

\section{Methodology: In-Distribution Module}
\label{section_in_distribution_module}
\begin{figure}[ht!]
    \centering
    \includegraphics[width=1\linewidth]{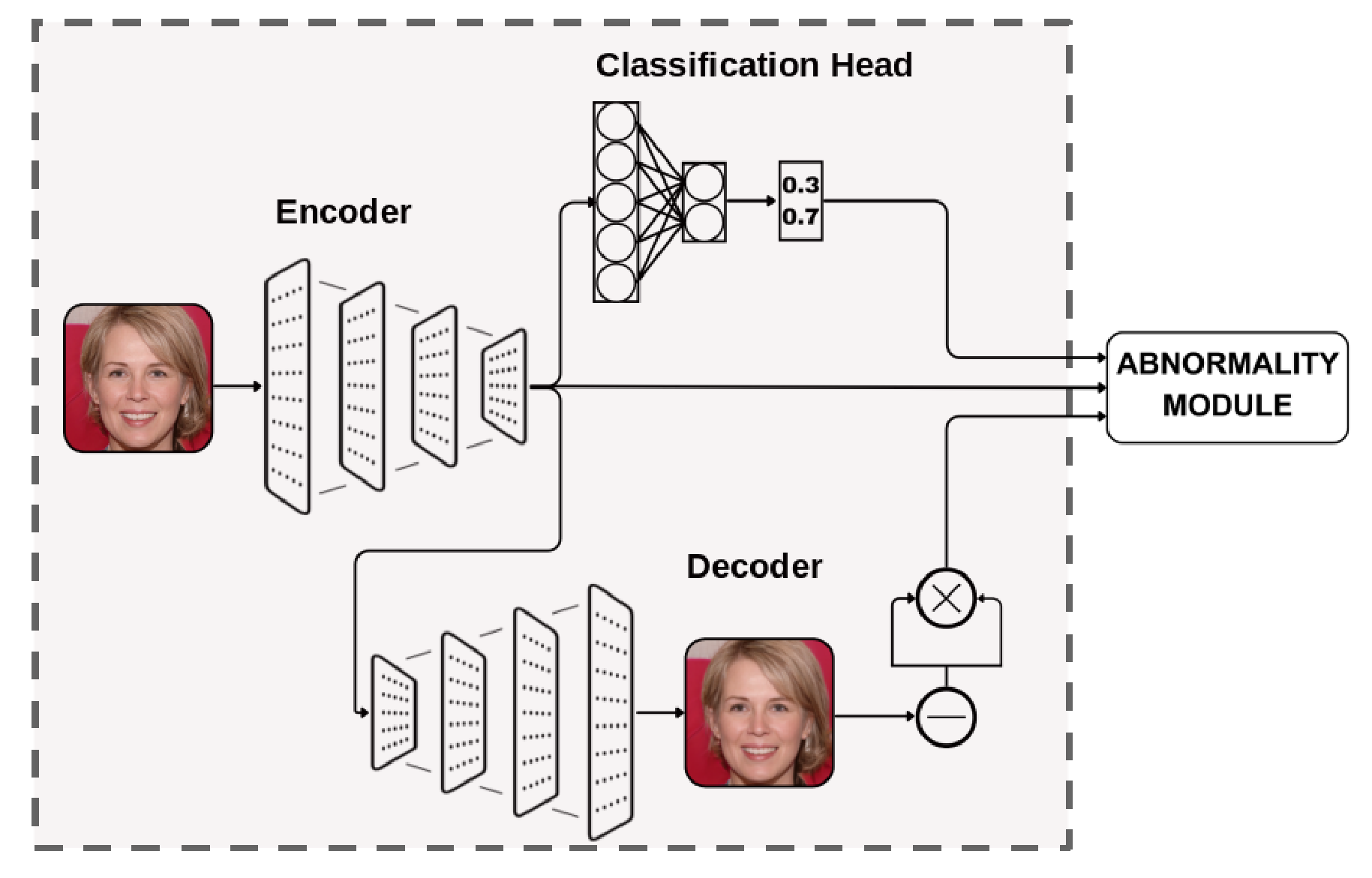}
    \caption{Pipeline architecture composed by ID Module and Abnormality Module.}
    \label{pipeline}
\end{figure}

The proposed methodology consists of two main modules. The first module (ID module) includes an encoder, a classification head, and a decoder. The decoder reconstructs input information, such as the image itself, and computes the squared residual. Similarly to Hendrycks and Gimpel~\cite{ood_baseline}, the residual, combined with the input encoding and the Softmax probability vector, is passed to the \emph{Abnormality} module. The Abnormality module processes this data stack to distinguish between ID and OOD samples. Figure \ref{pipeline} provides a detailed representation of this architecture, showcasing its application to a deepfake detection task, where it produces a classification vector for real and fake classes.
This section will introduce two variants of the ID module, one based on Convolutional networks and the other based on Transformer networks. Section~\ref{section_methodology} will instead introduce the OOD module.

In the remainder of this section, we present two variants of the ID module. One is based on convolutional networks, while the other is based on Transformer networks. Both networks share a common idea: the network is trained to learn to extract features that allow it to correctly classify the input as real or fake, and on the other hand, these features must contain enough information to reconstruct the original input. In this way, we ensure that the network learns to extract features that are descriptive of the training distribution. Furthermore, we expect the reconstruction error in the OOD module to increase when inputs come from unknown distributions.


\subsection{CNN-based Approach}
\label{methodology:id-cnn}
This ID module builds on the U-Net~\cite{model_unet} to identify deepfakes. U-Net is a fully convolutional network made of three main components: (i) the encoder, (ii) the bottleneck and (iii) the decoder. Encoder blocks start from the upper spatial dimension and scale, while decoder blocks start from the lower spatial dimension and scale. We choose this architecture because its nature of autoencoder allows us to obtain a compressed representation of the input image that we will later also use to identify OOD examples.

From now on with the term "layers", we refer to the number of Encoder and Decoder blocks and we use the level directly in the model name to specify this information, i.e. U-Net4 is composed of 4 Encoders and 4 Decoders. All the U-Net-based models respect two conditions. The first one concerns the number of features in the bottleneck, considering $l$ the U-Net layer, they are equal to $2^{4+l}$. The second one is about scaling the spatial dimension of the data after each layer, which is always by a factor of 2. So, a Decoder block reduces the dimension by half and the Encoder increases the spatial resolution up to double.
Considering these relationships it is easy to understand the architecture behind the custom networks U-Net4 and U-Net5. 

In order to identify deepfakes, we modify the original architecture by introducing a classification head after the bottleneck. This additional branch is composed of a sequence of five linear layers, interspersed with 1D batch normalizations, dropout layers with 30\% dropout probability, and RelU activation functions. We refer to this model as \emph{U-Net Scorer}.

We train the model with a weighted linear interpolation-like loss function reported in Equation \ref{loss_resnetEDS}. Symbols $x$ and $y$ represent the input sample and its label respectively, $\hat{x}$ is the reconstructed image, and $p$ is the classification prediction.
\begin{equation}
    \mathcal{L} = \alpha \, \mathcal{L}_{c} + \beta \, \mathcal{L}_{r} =  \alpha \, \text{BCE}(y,  p) + \beta \, \text{MSE}(x, \hat{x})
    \label{loss_resnetEDS}
\end{equation}

BCE represents the binary cross-entropy loss, while MSE is the Mean Squared Error loss, the classic function used for regression problems. In substitution of MSE it is possible to use the MAE loss.
Whether are respected the following conditions (\ref{lerp_cond}), the loss function becomes an exact linear interpolation (\ref{lerp}).
\begin{equation}
   \alpha + \beta = 1,\,\,\,\,\, 0 \leq \alpha \leq 1,\,\,\,\,\,  0 \leq \beta \leq 1   \\
   \label{lerp_cond}
\end{equation}
\begin{equation}
   \mathcal{L} = w \, \mathcal{L}_{c} + (1-w) \, \mathcal{L}_{r}
   \label{lerp}
\end{equation}

Taking into account that we want to model the classification as a task at a higher priority, and the reconstruction as a auxiliary target, are assigned the values $\alpha=0.9$ and $\beta=0.1$. Finally, we apply augmentation techniques like vertical/horizontal image flip as well as RandAugment~\cite{rand_aug}, which applies random transformations to the original image and CutMix technique~\cite{ood_cutmix}, which combines patches of different images to create augmented samples.

\subsection{Transformer-based Approach}
\label{methodology:id-transformer}
\begin{figure}[!h]
\centering
 \includegraphics[width=0.65\linewidth]{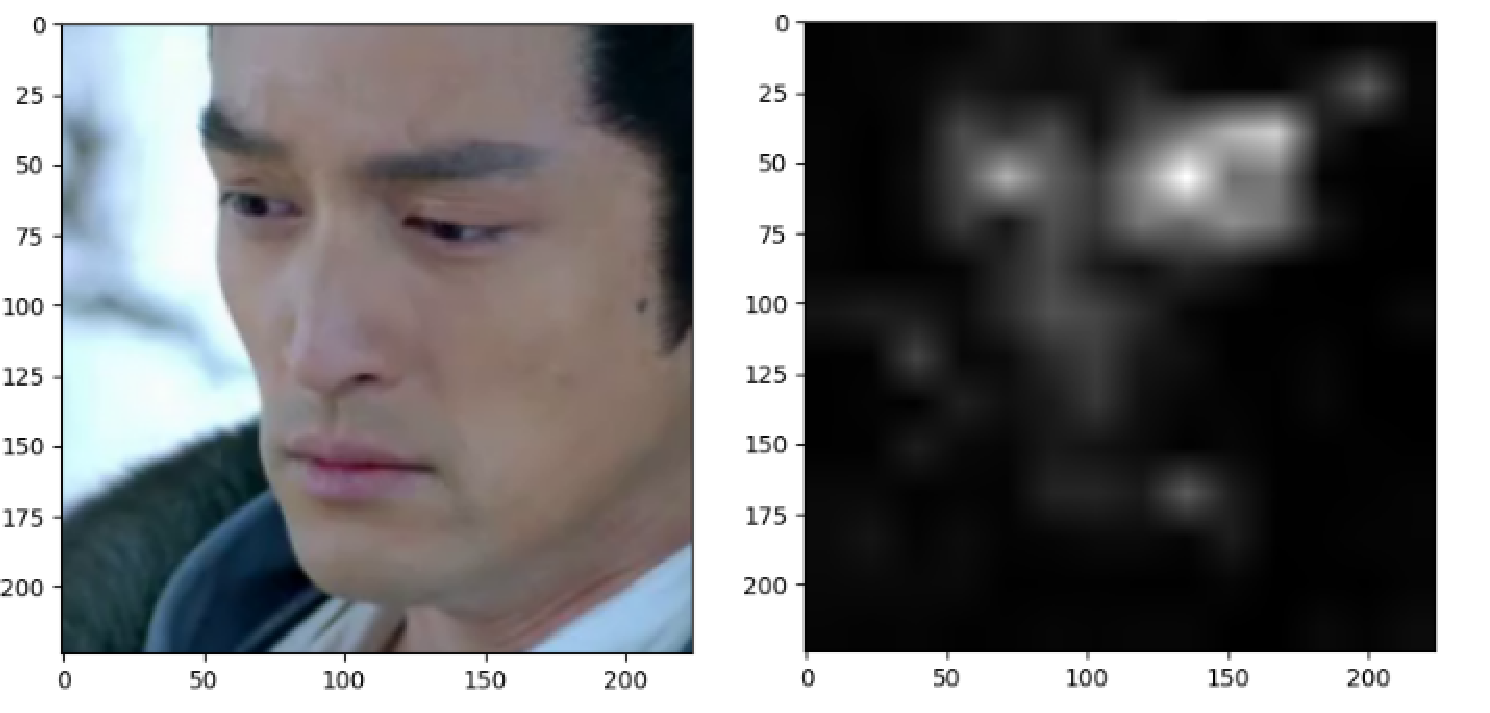}
  \caption{Attention heatmap from DeiT model.}
  \label{att_map_vit_fig}
\end{figure}

Building on the U-Net approach, we propose a novel method inspired by Cultrera et al. \cite{ood_vit}. Specifically, we utilize attention heatmaps generated by a ViT classifier as a means to detect Out-of-Distribution (OOD) samples. To achieve this, we design a convolutional network capable of reconstructing these attention heatmaps, enabling precise OOD detection without the need for additional supervision or labels during training. Instead of the original ViT, we use the Data-efficient Image Transformer (DeiT) to extract attention maps from the input images. These attention maps highlight the regions of the image the model focuses on, effectively capturing global context and long-range dependencies. By analyzing the distribution of these attention maps, the method detects Out-of-Distribution (OOD) samples. We then train an Autoencoder (AE) or Variational Autoencoder (VAE)~\cite{model_autoencoder} to reconstruct attention maps from In-Distribution (ID) samples. 
Table \ref{vae_t} shows the VAE model structure, with \textit{B} representing the batch size.
At test time, maps from OOD samples generally result in higher reconstruction errors, which are used to identify OOD data. 
This pipeline allows the model to effectively distinguish ID from OOD samples by focusing on the attention map reconstruction quality. An example of an extracted map is shown in Figure \ref{att_map_vit_fig}.

Unlike the U-Net-based solution, this approach trains the classifier and the reconstruction model separately. For the reconstruction process, two network architectures are proposed: the classic AutoEncoder (AE) and the Variational AutoEncoder (VAE). The AE is trained using the standard Mean Absolute Error (MAE) loss function. In contrast, the VAE employs a loss function that combines the MAE reconstruction loss with a KL divergence term, as shown in Equation (\ref{VAE_loss}). 
\begin{equation}
\mathcal{L}= \frac{1}{N} \sum_{i=1}^{N} |x_i - \hat{x}_i|  -0.5 \times \sum(1 + \log(\sigma^2) - \mu^2 - \sigma^2)
\label{VAE_loss}
\end{equation}
Here, \(N\) represents the total number of input components, \(x_i\) is an input sample, \(\hat{x}_i\) is its reconstruction, and \(\mu\) and \(\sigma\) are the mean and standard deviation used in the reparameterization trick.
Specifically, $N$ are all the input components, $x_i$ is the input sample, $\hat{x}_i$ represents the reconstruction, $\sigma$ and $\mu$ are respectively the mean and standard deviation for the reparametrization trick.

\begin{table}[!h]
\centering
\begin{tabular}{|l|l|c|}
\hline
\textbf{Layer/Module} & \textbf{Output Shape} & \textbf{Block} \\
\hline\hline
Input & (\textit{B}, 1, 224, 224) & \multirow{8}*{Encoder} \\
Conv2D + BN + Leaky ReLU& (\textit{B}, 32, 112, 112) &  \\
Conv2D + BN + Leaky ReLU& (\textit{B}, 64, 56, 56) &  \\
Conv2D + BN + Leaky ReLU& (\textit{B}, 128, 28, 28) &  \\
Conv2D + BN + Leaky ReLU& (\textit{B}, 256, 14, 14) &  \\
Conv2D + BN + Leaky ReLU& (\textit{B}, 512, 7, 7) &  \\
Flatten & (\textit{B}, 25088) &  \\
Linear & (\textit{B}, 1024) &  \\
\hline\hline
Linear + Leaky ReLU & (\textit{B}, 25088) & \multirow{8}*{Decoder} \\
Unflatten & (\textit{B}, 512, 7, 7) &  \\
ConvTranspose2D + BN + Leaky ReLU& (\textit{B}, 256, 14, 14) &  \\
ConvTranspose2D + BN + Leaky ReLU & (\textit{B}, 128, 28, 28) &  \\
ConvTranspose2D + BN + Leaky ReLU& (\textit{B}, 64, 56, 56) &  \\
ConvTranspose2D + BN + Leaky ReLU & (\textit{B}, 32, 112, 112) &  \\
ConvTranspose2D + Sigmoid & (\textit{B}, 1, 224, 224) &  \\ \hline
\end{tabular}
\caption{Variational Autoencoder architecture.}
\label{vae_t}
\end{table}

\section{Out-Of-Distribution Module}
\label{section_methodology}
We now describe the Out-Of-Distribution (OOD) module. This detector enhances its ability to identify OOD samples by utilizing diverse data generated by the In-Distribution (ID) module. Our approach, which we call the \emph{Abnormality module}, builds on the foundational concept of the Softmax baseline introduced by Hendrycks and Gimpe~\cite{ood_baseline}. This module is designed to recognize instances that deviate from the norm or are underrepresented in the training data, thereby improving the OOD detection capabilities of machine learning models.

The Abnormality module utilizes all the critical information learned during the training of the deepfake detector. It processes the model's raw output values, called logits, and converts them into probabilities using the Softmax activation function. This transformation creates a probability distribution that reflects the model's confidence in its predictions. However, while these confidence scores indicate how certain the model is about each prediction, they are not always sufficient to clearly distinguish between In-Distribution (ID) samples (familiar data from training) and Out-Of-Distribution (OOD) samples (new, unseen data). Typically, confident predictions result in probabilities that are either close to 0 (indicating low likelihood) or close to 1 (indicating high likelihood). Despite this, confidence scores are only one component of a more comprehensive detection framework designed to improve OOD detection.

Another crucial component is the reconstruction residual, which evaluates the decoder's ability to reconstruct input data. This ability decreases when the decoder encounters samples from outside the known distribution, providing a strong indicator for OOD detection. Additionally, the encoding offers a latent representation of the input from the classifier's perspective, acting as an effective discriminator for significantly different samples. The calculation of the reconstruction residual is tailored to fit the specific architecture of the ID module, ensuring both compatibility and optimal performance.

\subsection{CNN-based Approach}
In the CNN-based approach (introduced in Section~\ref{methodology:id-cnn}), image reconstruction is used to compute the residual by squaring the difference between the reconstructed image and the original input. Unless specified otherwise, the image encoding is always extracted at the bottleneck layer.
To manage the higher dimensionality of the encoding, it is recommended to use the U-Net 4 Scorer for images sized $112\times112$, and the U-Net 5 Scorer for images sized $224\times224$. Alternatively, a smaller encoding can be extracted from the layers of the classification head. While this is not strictly necessary, using an oversized encoding can lead to issues in the Abnormality module, such as data imbalance and longer convergence times.

\subsection{Transformer-based Approach}
As described in Section~\ref{methodology:id-transformer}, the Transformer-based approach replaces image residual computation with attention map residual computation. Unlike the CNN-based method, it calculates the residual from a grayscale heatmap and its reconstruction via a Variational Autoencoder (VAE), which is then squared. This auxiliary data, along with the latent image representation and Softmax probabilities from the classification head, is used as input for the Abnormality module.
The image encoding is extracted from the Transformer encoder before the classification MLP block, specifically using features associated with the [CLS] token. As in the previous approach, the probability vector is computed using the Softmax function.

\subsection{Abnormality Modules}
The Abnormality module is built on the Softmax baseline~\cite{ood_baseline}, which serves as a simple and fundamental solution for this architecture. The auxiliary data from the ID module is fed into the Abnormality module, where the residual is flattened, and all auxiliary data is combined into a single tensor. This combined data is then processed through a Multilayer Perceptron (MLP). The final output logit is passed through a sigmoid activation function to produce the \emph{risk} value, which indicates the likelihood of the sample being Out-Of-Distribution (OOD).

Given $l$ as the U-Net level, $R$ as the image resolution (assumed to be square, with equal width and height), and $C$ as the number of input channels, the following sizes can generally be computed using these parameters.

\begin{equation}
    \text{\# Encoding (U-Net)} = R^2 \cdot 2^{4-l}\newline
\end{equation}
\begin{equation}
    \text{\# Residual (flatten)} = R^2 \cdot C
\end{equation}
The large dimensionality resulting from the concatenation in this model leads to high computational costs and slow convergence. Additionally, the significant size differences among the various auxiliary data types can overshadow their individual contributions. This issue particularly affects the Softmax probability tensor, which is the most impacted by this imbalance.

\begin{figure}[]
    \centering
    \includegraphics[width=1\linewidth]{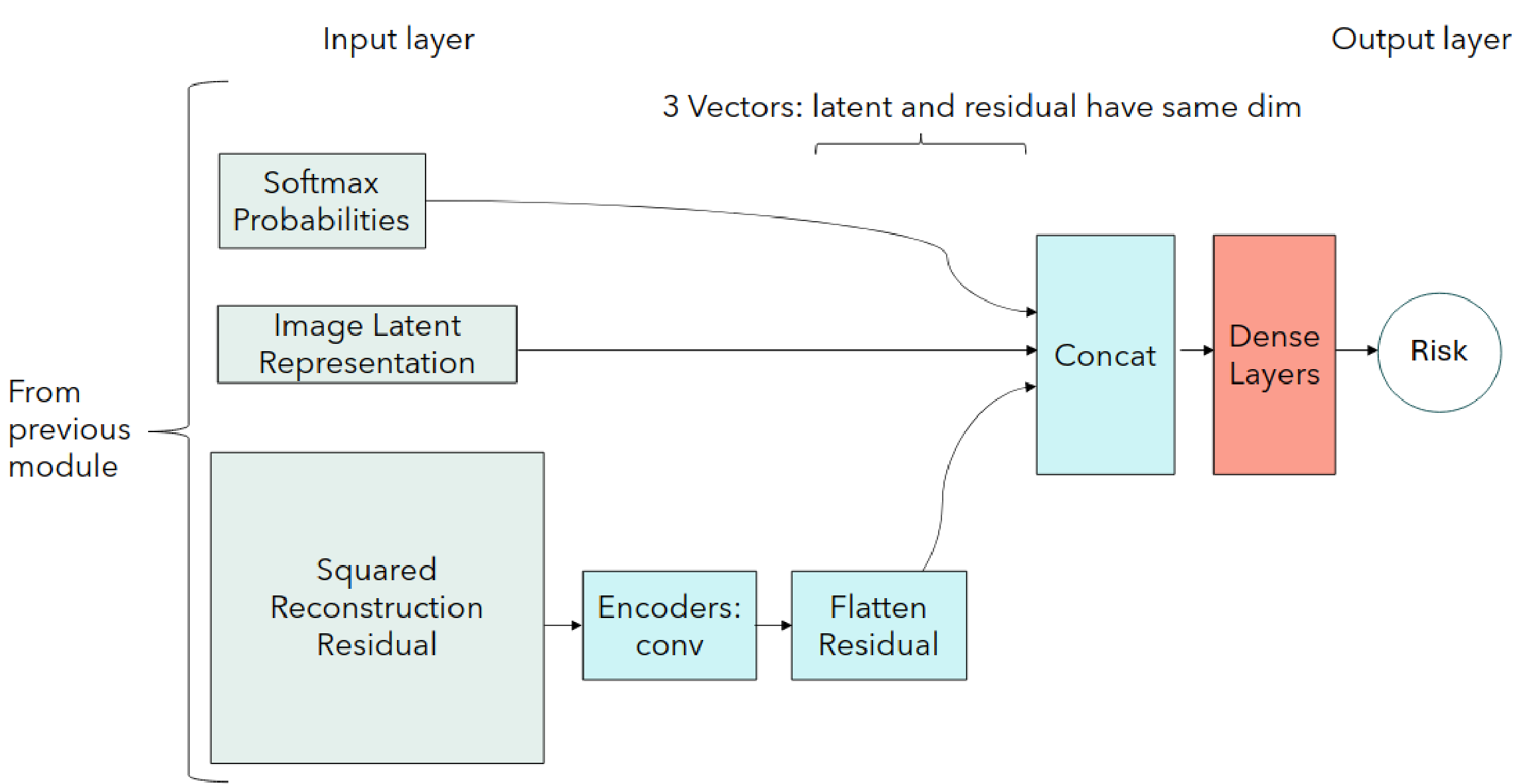}
    \caption{Abnormality module Encoder V1 and V2 scheme.}
    \label{abn_enc_v1v2_s}
\end{figure}

We propose three different Abnormality module architectures ($V1$, $V2$, and $V3$) that extend the basic version by incorporating additional encoders. These enhancements aim to make the model more efficient and improve performance by balancing the data.
In this study, the dimensions are based on using a U-Net4 Scorer as the ID module with input images of 112p resolution. The first variant, $V1$, introduces a slight modification to the basic Abnormality module by reducing the residual size before concatenation with other auxiliary data. This is achieved through a new sub-module called the Encoder block, designed to reduce spatial dimensions while increasing the number of features.
The Encoder block consists of a sequence of three components: a 2D Convolutional layer, 2D Batch Normalization, and a GELU activation function. The second convolutional layer in the sequence performs downsampling with a stride of 2, while each block increases the number of features from the initial convolutional layer. Three such encoding blocks are used to reshape the residual information, followed by concatenation with other data and processing through the MLP block, as described earlier. The full structure is illustrated in Figure~\ref{abn_enc_v1v2_s} and detailed in Table~\ref{abn_enc_v1_t}.
By using this encoding in the Abnormality module, the residual information is reduced to the same dimension as the ID classifier encoding. While the imbalance with Softmax probabilities remains, the imbalance between encoding and residual is resolved. As a result, computational complexity and training time are significantly reduced.

\begin{table}[]
\centering
\begin{tabular}{ |l|l|c|}
\hline\
\textbf{Layer/Module} & \textbf{Output Shape} & \textbf{Branch} \\ \hline\hline
Input 1 & (\emph{B}, 2) & Softmax \\  \hline\hline
Input 2 & (\emph{B}, 12544) & Encoding \\ \hline\hline
Input 3 & (\emph{B}, 3, 112, 112) & \multirow{5}*{Residual} \\ 
Encoder block 1& (\emph{B}, 16, 56, 56) &  \\
Encoder block 2& (\emph{B}, 32, 28, 28) &  \\
Encoder block 3& (\emph{B}, 64, 14, 14) &  \\
Flatten & (\emph{B}, 12544) &  \\ \hline\hline
Concatenation & (\emph{B}, 25090) & \multirow{7}*{Risk} \\
FC 1 + BN + GELU& (\emph{B}, 4096) &  \\
FC 2 + BN + GELU& (\emph{B}, 2048) &  \\
FC 3 + BN + GELU& (\emph{B}, 1024) &  \\
FC 4 + BN + GELU& (\emph{B}, 512) &  \\
FC 5 + BN + GELU& (\emph{B}, 128) &  \\
FC Final + Sigmoid & (\emph{B}, 1) &  \\ \hline
\end{tabular}
\caption{Abnormality module Encoder V1 architecture table.}
\label{abn_enc_v1_t}
\end{table}

The advanced version, $V2$, further enhances residual encoding by adjusting the encoding extraction without adding additional downsampling blocks. Instead of extracting the encoding from the bottleneck as in previous versions, it is now taken directly from the first convolutional layer in the classification or scorer branch of the U-Net Scorer model or from a smaller embedding size in the case of a ViT. This modification comes at no extra computational cost.
An additional encoding block is included in the residual branch to ensure the encoding and residual have matching dimensions. The overall architecture remains the same as shown in Figure~\ref{abn_enc_v1v2_s}, with the specific layers and internal dimensionalities updated as detailed in Table~\ref{abn_enc_v2_t}.

\begin{table}[]
\centering
\begin{tabular}{|l|l|c|}
\hline
\textbf{Layer/Module} & \textbf{Output Shape} & \textbf{Branch} \\
\hline\hline
Input 1 & (\emph{B}, 2) & Softmax \\
\hline\hline
Input 2 & (\emph{B}, 1568) & Encoding \\
\hline\hline
Input 3 & (\emph{B}, 3, 112, 112) & \multirow{6}*{Residual} \\
Encoder block 1 & (\emph{B}, 4, 56, 56) &  \\
Encoder block 2 & (\emph{B}, 8, 28, 28) &  \\
Encoder block 3 & (\emph{B}, 16, 14, 14) &  \\
Encoder block 4 & (\emph{B}, 32, 7, 7) &  \\
Flatten & (\emph{B}, 1568) &  \\
\hline\hline
Concatenation & (\emph{B}, 3138) & \multirow{5}*{Risk} \\
FC 1 + BN + GELU& (\emph{B}, 1024) &  \\
FC 2 + BN + GELU& (\emph{B}, 512) &  \\
FC 3 + BN + GELU& (\emph{B}, 128) &  \\
FC Final + Sigmoid & (\emph{B}, 1) &  \\
\hline
\end{tabular}
\caption{Abnormality module Encoder V2 architecture table.}
\label{abn_enc_v2_t}
\end{table}

\begin{figure}[!ht]
    \centering
    \includegraphics[width=0.9\linewidth]{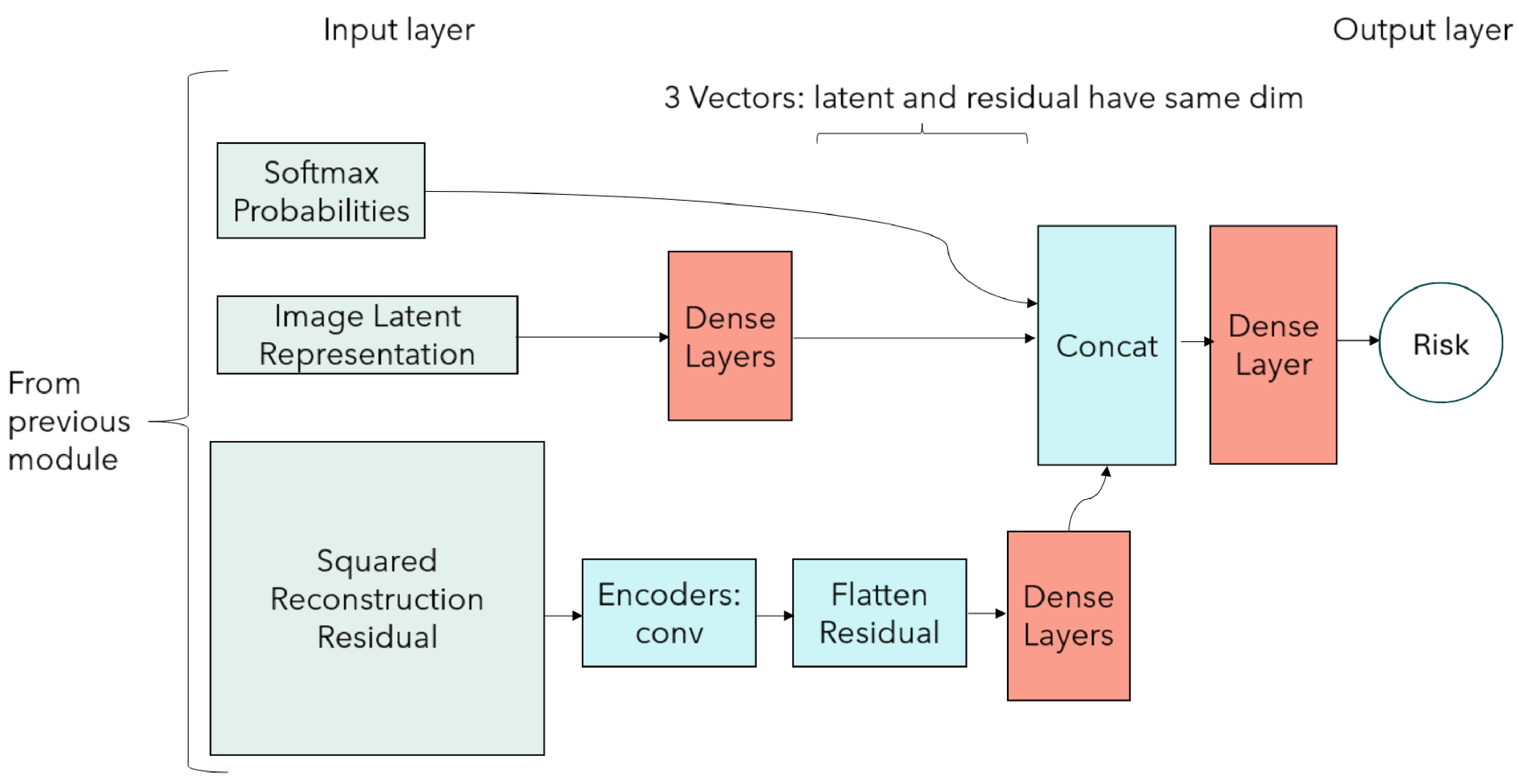}
    \caption{Abnormality module Encoder V3 scheme.}
    \label{abn_enc_v3_s}
\end{figure}

The final version, $V3$, further improves the concept of balancing auxiliary data. This is achieved by introducing two MLP blocks, one for the Encoding branch and one for the Residual branch. Starting with the structure of Abnormality module $V1$, which uses a large encoding input and three Encoding blocks for the residual, these new MLP blocks have a consistent design featuring five fully connected layers, batch normalization, and GELU activation.
In contrast, the risk MLP block is simplified, consisting of a single dense layer followed by a sigmoid activation function. Figure~\ref{abn_enc_v3_s} and Table~\ref{abn_enc_v3_t} provide an overview and additional details of this architecture.

\begin{table}[]
\centering
\begin{tabular}{|l|l|c|}
\hline
\textbf{Layer/Module} & \textbf{Output Shape} & \textbf{Branch} \\
\hline\hline
Input 1 & (\emph{B}, 2) & Softmax \\
\hline\hline
Input 3 & (\emph{B}, 3, 112, 112) & \multirow{11}*{Residual} \\
Encoder block 1& (\emph{B}, 16, 56, 56) &  \\
Encoder block 2 & (\emph{B}, 32, 28, 28) &  \\
Encoder block 3 & (\emph{B}, 64, 14, 14) &  \\
Flatten & (\emph{B}, 12544) &  \\
FC R1 + BN + GELU& (\emph{B}, 4096) &  \\
FC R2 + BN + GELU& (\emph{B}, 2048) &  \\
FC R3 + BN + GELU& (\emph{B}, 1024) &  \\
FC R4 + BN + GELU& (\emph{B}, 512) &  \\
FC R5 + BN + GELU& (\emph{B}, 128) &  \\
FC R6 + BN + GELU& (\emph{B}, 10) &  \\
\hline\hline
Input 2 & (\emph{B} , 12544) & \multirow{7}*{Encoding} \\
FC E1 + BN + GELU& (\emph{B}, 4096) &  \\
FC E2 + BN + GELU& (\emph{B}, 2048) &  \\
FC E3 + BN + GELU& (\emph{B}, 1024) &  \\
FC E4 + BN + GELU& (\emph{B}, 512) &  \\
FC E5  + BN + GELU& (\emph{B}, 128) &  \\
FC E6 + BN + GELU& (\emph{B}, 10) &  \\
\hline\hline
Concatenation & (\emph{B}, 22) & \multirow{2}*{Risk} \\
FC Final + Sigmoid & (\emph{B}, 1) &  \\
\hline
\end{tabular}
\caption{Abnormality module Encoder V3 architecture table.}
\label{abn_enc_v3_t}
\end{table}

The significant reduction of both encoding and residual to just 10 elements before concatenation is designed to achieve a strong balance between the three inputs. It's important to note that this applies to all the Abnormality modules discussed. Specifically, when using higher-resolution images (e.g., switching from 112p to 224p), it is necessary to add an additional encoder block to the residual branch to maintain consistent tensor shapes. The encoding dimension remains the same, and this holds true whether the initial assumption of using 112p images with a U-Net4-like model or 224p images with a U-Net5-like model is followed.
These proposed models are trained using the standard Binary Cross-Entropy (BCE) loss function.

These 3 abnormality module versions are all valid for the CNN approach, and can be employed with both 112p and 224p resolution images. In the case of the Transformer approach instead, only V2 and V3 architectures are appropriate..\\
The tiny DeiT as ID classifier works only with 224p images, and it produces a small size encoding of 192 elements. Given this reduced latent representation, the V1 solution designed with a large encoding size, cannot be used for this case. Therefore, V2 and V3 are the solutions that best suit the Transformer use, mitigating as possible any related problem on auxiliary data balancing. In this case, the equality in size between the encoding and residual information in the concatenation phase is not respected, but the relatively small difference between those validates the process.

\subsection{Abnormality Modules Training}
\begin{figure}[]
    \centering
    \includegraphics[width=1\linewidth]{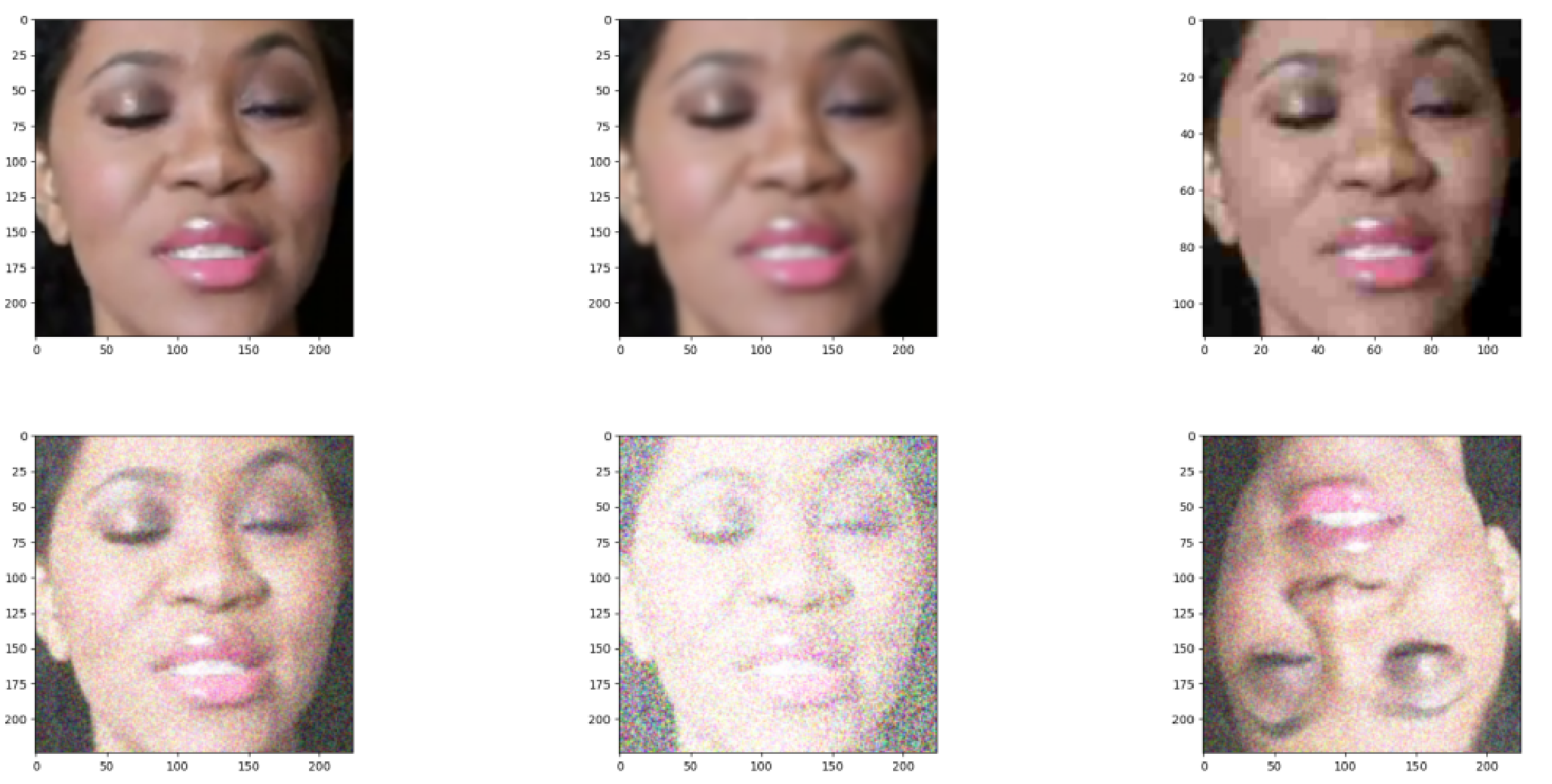}
    \caption{Alteration techniques for synthesized OOD data.}
    \caption*{From left to right and top to bottom:  original sample, blur effect, JPEG compression, normal noise, normal noise distorted, normal noise, and random rotation}
    \label{synthesized_ood}
\end{figure}

A crucial aspect to consider, concerns data applied to train the OOD detector. Differently from ID module, both ID and OOD data are necessary to train the Abnormality module, and we can distinguish between two training methodologies that can be exploited. Both solutions consider the ID train set as the one employed by the classifier, but they differ in the nature of OOD data. The first methodology uses five techniques to synthesize OOD samples, altering the ID images. Specifically: Blurring from Gaussian filter, JPEG compression at high loss (10\% of quality), inclusion of normal noise and normal noise with random standard deviation (distortion effect), and inclusion of noise with $\frac{\pi}{2}$ random image rotation. Examples of these techniques are presented in Figure \ref{synthesized_ood}. Using these data, we exploit an unsupervised technique to train the Abnormality module. In contrast, The second methodology integrates available OOD samples with the role of real outliers, defining a supervised procedure. In addition, it's also possible to define a third possibility in which these two methodologies are mixed together.

\section{Results}\label{section_results}
\subsection{Deepfake Dataset and Preprocessing}
Our work, proposed a novel usage of CDDB (Continual Deepfake Detection Benchmark) Dataset \cite{cddb} by Huang et al. adapting the full dataset to the requirements of our task. As its name suggests this dataset provides a benchmark designed to evaluate deepfake detection methods in a continual learning setting. It offers a realistic and challenging environment for testing deepfake detection algorithms. CDDB includes 11 deepfake models divided into 3 different groups by source: GAN models, non-GAN models, and unknown models, for an overall number of images equal to 82k. The main goal of this data processing step is to transform a continual learning dataset, with different binary classifications, into a single classification dataset OOD oriented. For the pure Deepfake detection task, the binary and binary-partial CDDBs are designed, the first includes the whole set of data available, while the second proposes a division between In-Distribution and Out-Of-Distribution. ID and OOD split is necessary to effectively face Deepfake and OOD detection in the same work pipeline. This data separation is thought to follow three different strategies. In the \textit{content scenario}, the semantic content of the data (faces/not faces) is the discriminant to move data in the In or Out distribution set. For each Deepfake technique, the data in most cases is well separated by the image’s subject. The first study case that we define, uses images containing faces as ID and the rest of the set as OOD. It’s important to notice that several DF techniques are not organized by content but mix the presence of different subjects. To avoid semantic overlapping between ID and OOD these sets have been removed from this study. The \textit{group scenario} uses as key to separate samples, the DF methods groups. Our choice to define the relative study case, is to use GAN models as ID data and the rest as OOD data. The last scenario defined, called the \textit{mix scenario}, separates samples based on techniques, assigning randomly to each Deepfake method either the ID or the OOD set. The assignment even if random, is capable of balancing the two partitions to avoid insufficient data for the Deepfake classifier. In the specific for the ID data we have: BigGAN, GauGAN, StarGAN, Deepfake dataset, Glow, CRN, and WildDeepFake sets. The other pre-processing operations include the definition of a validation set sampled from the test data, downsampling/upsampling to a fixed resolution, and normalization of images.

\subsection{Deepfake Detection}
Following the architecture described for the two approaches as first are exposed the evaluation performances for the CNN-based approach. We introduce the metrics of U-Net4 Scorer trained with 112p images. These results are then compared with the ones from the U-Net5 Scorer trained with 224p images. With this comparison we want to demonstrate how the downsampling of input information could be critical, removing key information useful for correct inferences. U-Net4 Scorer with 112p input images and U-Net5 Scorer with 224p images, have the same memory impact of around 2,760 MB. Things change whether we apply $224 \times 224$ images with U-Net4 Scorer, incrementing the latent encoding size processed from the classification head, leading to a memory request equal to 3,602 MB. Motivated by this 30\% memory increase, it’s suggested the utilization of U-Net5 Scorer with the standard 224 pixels of spatial dimensions. Table \ref{metrics_unets} gives a complete comparison of the metrics.\\
\begin{table}[!h]
    \centering
    \begin{tabular}{|l|c|c|}
        \hline
        \textbf{Metric} & \textbf{U-Net4 Scorer} & \textbf{U-Net5 Scorer} \\ \hline
        Accuracy & 0.906 & 0.946 \\ 
        Precision & 0.904 & 0.957 \\
        Recall & 0.910 & 0.934 \\ 
        F1-Score & 0.907 & 0.945 \\
        AUROC & 0.968 & 0.991 \\
        AUPR & 0.968 & 0.991 \\
        Jaccard Score & 0.881 & 0.897 \\ \hline
    \end{tabular}
    \caption{Performance Metrics for U-Net Scorer, Content scenario.}
    \label{metrics_unets}
\end{table}
 \textit{Group} and  \textit{Mix scenarios} have proven to be more complex for detection. In Table \ref{unet5_table_gan_mix} the metrics are summarized using U-Net5 Scorer.
\begin{table}[!h]
    \centering
    \begin{tabular}{|l|c|c|}
        \hline
        \textbf{Metric} & \textbf{Group scenario} & \textbf{Mix Scenario} \\ \hline
        Accuracy    & 0.906 & 0.926 \\ 
        Precision   & 0.902 & 0.950 \\ 
        Recall   & 0.914 & 0.900 \\ 
        F1-Score & 0.908 & 0.924  \\ 
        AUROC & 0.973 & 0.988 \\
        AUPR & 0.973 & 0.988 \\ 
        Jaccard Score & 0.831 & 0.860 \\ \hline
    \end{tabular}
    \caption{Performance Metrics for U-Net5 Scorer, over  \textit{Group} and  \textit{Mix scenarios}.}
    \label{unet5_table_gan_mix}
\end{table}
We conclude this section on the results of pure deepfake detection exposing the performances of the Transformer-based solution which involves the employment of the tiny DeiT model for the detection. The evaluation is proposed for the Content scenario. This model has a reduced memory impact if compared to the previous model, with only 1,118 MB of video memory necessary. This leads to a reduction of almost 57\% of memory usage with respect to the 2,760 MB from U-Net5 Score. This model reaches an accuracy and precision of 0.96, with a recall of about 0.956. The resulting F1-score corresponds to 0.96. Moreover, the Jaccard score is 0.92, with AUROC and AUPR both 0.993. 

\subsection{Out-Of-Distribution Detection}
Based on the results obtained in the deepfake detection exploiting the different approaches and models, it’s possible to discuss the Out-Of-Distribution detection performances of the proposed work, We start with results coming from different methods in the literature under the three different scenarios defined in the CDDB section: \textit{Content}, \textit{Group}, and \textit{Mix}. The baselines proposed are the MSP \cite{ood_baseline}, the ODIN \cite{ood_odin}, and the Confidence branch \cite{ood_conf_branch} for the Content scenario. The confidence branch solution employs U-Net4 Scorer for 112p images as ID classifier, while both MSP and ODIN use U-Net5 Scorer upstream with 224p images.  Follows the Tables \ref{baselines_content}, \ref{baselines_group}, \ref{baselines_mix} for each scenario.\\
\begin{table}[!h]
    \centering
    \begin{tabular}{|l|c|c|c|}
        \hline
        \textbf{Metric} & \textbf{MSP} & \textbf{ODIN} & \textbf{Conf. branch}\\ \hline
        AUROC    & 77.85 & 82.67 & 78.38\\ 
        AUPR   & 71.61 & 80.03 & 68.17\\ 
        FPR95   & 0.61 & 0.56 & 0.61\\   \hline
    \end{tabular}
    \caption{Baselines performances on \textit{Content scenario}.}
    \label{baselines_content}
\end{table}
\begin{table}[!h]
    \centering
    \begin{tabular}{|l|c|c|}
        \hline
        \textbf{Metric} & \textbf{MSP} & \textbf{ODIN} \\ \hline
        AUROC    &  62.06 & 60.3 \\
        AUPR   & 58.9 & 54.31\\
        FPR95   & 0.84 & 0.83 \\  \hline
    \end{tabular}
    \caption{Baselines performances on \textit{Group scenario}.}
    \label{baselines_group}
\end{table}
\begin{table}[!h]
    \centering
    \begin{tabular}{|l|c|c|}
        \hline
        \textbf{Metric} & \textbf{MSP} & \textbf{ODIN} \\ \hline
        AUROC    & 69.74 & 62.58 \\ 
        AUPR   & 61.6 & 55.05\\
        FPR95   & 0.74 & 0.823 \\ \hline
    \end{tabular}
    \caption{Baselines performances on \textit{Mix scenario}.}
    \label{baselines_mix}
\end{table}
\noindent
At this point, it's possible to compare these baselines with our three abnormality proposed modules. Each module can be trained with synthesized OOD data, real OOD samples, or mixing them. The following acronyms are used to indicate this specific: \textit{S} for synthesized, \textit{R} for real, and \textit{M} for mixed data.
The first comparison we propose is between the three Abnormality Modules, trained and tested on the Content scenario with U-Net4 Scorer as ID classifier. The classifier accuracy is equal to 0.906 and all the OOD metrics are expressed in Table \ref{abns_gan_faces}.
\begin{table}[!h]
    \centering
    \begin{tabular}{|l|c|c|c|c|c|c|}
        \hline
        \textbf{Metric} & \textbf{V1S} & \textbf{V2S} & \textbf{V3S} & \textbf{V1R} & \textbf{V2R} & \textbf{V3R} \\ \hline
        AUROC  &77.79& 91.6 &64.13&99.97& 100&99.89\\ 
        AUPR   &80.06& 93.55 &66.65&99.96& 100&99.9\\
        FPR95  &0.83& 0.49 &0.94&2e-4& 0&2e-4\\ \hline
    \end{tabular}
    
\vspace{5px}

    \begin{tabular}{|l|c|c|c|}
        \hline
        \textbf{Metric} &\textbf{V1M} & \textbf{V2M} & \textbf{V3M} \\ \hline
        AUROC  &99.93&99.95&97.45\\ 
        AUPR   &99.91&99.88&95.98\\ 
        FPR95  &08e-4&8e-4&0.04\\ \hline
    \end{tabular}
    \caption{Custom CNN Abnormality modules performances on \textit{Content scenario}.}
    \label{abns_gan_faces}
\end{table}
\noindent
The most interesting outcome is the one from V2S since produces good results using only altered ID data treated as OOD samples in the training phase, providing an interesting unsupervised approach. Moving toward a more complex task, other results on the Group and Mix scenario are proposed in Table \ref{abns_gan_group_mix}. This time the ID classifier is the U-Net5 Scorer which produces an accuracy of 0.906 for the Group and 0.926 for the Mix scenario. The results are exposed only for the module that better performs, which as demonstrated is V2.
\begin{table}[!h]
    \centering
    \begin{tabular}{|l|c|c|c|c|}
        \hline
        \textbf{Metric} & \textbf{V2S Group} & \textbf{V2M Group} & \textbf{V2S Mix} & \textbf{V2M Mix} \\ \hline
        AUROC  & 63.35& 92.72& 60.14 & 80.0\\
        AUPR   & 64.21& 95.1& 65.48 & 86.19\\
        FPR95  & 0.90& 0.59& 0.94 &  0.87\\ \hline
    \end{tabular}
    \caption{Custom CNN Abnormality modules performances on  \textit{Group} and  \textit{Mix scenario}.}
    \label{abns_gan_group_mix}
\end{table}
It is clear that these scenarios are truly complex, despite this, good results are obtained by using real outliers.\\
Afterward the demonstration of results for the CNN-based approach, we move to evaluate how the Transformer-based technique performs. This approach requires an additional model for the reconstruction of the attention maps. In this case, a VAE model with a reconstruction MSE of $2e-4$ is included. In a similar manner to what has been shown for the CNN approach in the  \textit{Content scenario}, it’s proposed the same evaluation environment with this different inference pipeline for Abnormality modules V2 and V3. All data are inserted in the Table \ref{abns_deit_faces}.

\begin{table}[!h]
    \centering
    \begin{tabular}{|l|c|c|c|c|c|c|}
        \hline
        \textbf{Metric} & \textbf{V2S} & \textbf{V3S} & \textbf{V2R} & \textbf{V3R} & \textbf{V2M} & \textbf{V3M} \\ \hline
        AUROC  & 81.96& 93.77& 100&99.98&99.97&99.94\\
        AUPR   & 85.23& 94.55& 100&99.98&99.98&99.92\\
        FPR95  & 0.81& 0.41& 1e-4&2e-4&3e-4&3e-4\\ \hline
    \end{tabular}
    \caption{Custom Transformer Abnormality modules V1 and V2 performances on  \textit{Content scenario}.}
    \label{abns_deit_faces}
\end{table}
In this case, V3 performs surprisingly better than V2 module beating the best result of the U-Net approach on the training with only synthetic data. Considering synthesized OOD data, the Transformer approach overcomes the CNN one, while using real outliers available the performances are practically at the same level.
Tests and evaluations of the transformer approach over Group and Mix scenarios have highlighted the limit of this approach. The limitation is caused by the model's tendency to consider semantic information more highly than fundamental features that are essential for determining authenticity. As a consequence attention maps tend to become uniform during training whether images contain very varied content, failing to provide meaningful insights. Tests to prove this claim are conducted using larger configurations too, including DeiT base.

\subsection{Out-Of-Distribution Benchmark}
In this section are presented the results of our technique applied to CIFAR10 benchmark from \cite{openood}, expressed in Table \ref{benchmark_AUROC_far} and Table \ref{benchmark_AUROC_near}.\\
Each table is divided into three blocks: the first represents the Post-hoc inference methods which don’t require any training, the second block is related to trained methods without the necessity of OOD samples, while the third includes techniques that require OOD data as outliers. Our methods are named "ABN Module V2S" and "ABN Module V2M". The approach involved is the Transformer-based, reaching an ID accuracy of 97.5\%. As before, the name V2S represents the Abnormality module V2 trained using only synthesized OOD data, while V2M refers to the same module with a mix of synthetic and real OOD samples. Real OOD are data sampled from near OOD datasets as outliers. Thus, V2S belongs to the second block, while V2R is in the third one. Moreover, the evaluations are divided between Near and Far OOD, an empiric separation proposed by the benchmark itself.
In details for Far-OOD are assumed: MNIST, SVHN, and Texture datasets. While in Near-OOD are included the CIFAR100 and TinyImageNet.
Respectively, in Table \ref{benchmark_AUROC_far} and \ref{benchmark_AUROC_near}, we have AUROC metric for Far and Near OOD. Furthermore, for the other techniques, the tables present whether is available the average number of the metrics obtained from 3 training runs.

\begin{table}[hbtp]
    \footnotesize
    \centering
    \begin{center}
        \begin{tabular}{ | l | c | c | c  | c |}
        \hline
        \textbf{Technique} & \textbf{MNIST} & \textbf{SVHN} & \textbf{Textures} & \textbf{ID ACC} \\
        \hline\hline
        \multicolumn{5}{| c |}{\cellcolor{blue!15}\textbf{Post-Hoc}}\\ 
        \hline\hline
        OpenMax & 90.50 & 89.77 & 89.58 & 95.06 \\
        MSP & 92.63 & 91.46 & 89.89 & 95.06 \\
        TempScale & 93.11 & 91.66 & 90.01 & 95.06 \\
        ODIN & 95.24 & 84.58 & 86.94 & 95.06 \\
        MDS & 90.10 & 91.18 & 92.69 & 95.06 \\
        MDSEns & \underline{99.17} & 66.56 & 77.40 & 95.06 \\
        RMDS & 93.22 & 91.84 & 92.23 & 95.06 \\
        Gram & 72.64 & 91.52 & 62.34 & 95.06 \\
        EBO & 94.32 & 91.79 & 89.47 & 95.06 \\
        OpenGAN & 56.14 & 52.81 & 56.14  & 95.06 \\
        GradNorm & 63.72 & 53.91 & 52.07 & 95.06 \\
        ReAct & 92.81 & 89.12 & 89.38 & 95.06 \\
        MLS & 94.15 & 91.69 & 89.41 & 95.06 \\
        KLM & 85.00 & 84.99 & 82.35 & 95.06 \\
        VIM & 94.76 & 94.50 & 95.15 & 95.06 \\
        KNN & 94.26 & 92.67 & 93.16 & 95.06 \\
        DICE & 90.37 & 90.02 & 81.86 & 95.06 \\
        RankFeat & 75.87 & 68.15 & 73.46 & 95.06 \\
        ASH & 83.16 & 73.46 & 77.45 & 95.06 \\
        SHE & 90.43 & 86.38 & 81.57 & 95.06 \\
        GEN & 93.83 & 91.97 & 90.14 & 95.06 \\
        \hline\hline
        \multicolumn{5}{| c |}{\cellcolor{blue!15}\textbf{Trained}}\\ 
        \hline\hline
        ConfBranch & 94.49 & 95.42 & 91.10 & 94.88 \\
        RotPred & 97.52 & 98.89 & 97.30 & 95.35 \\
        RotPred + PixMix & 98.00 & \underline{99.33} & \underline{99.21} & 95.91 \\
        G-ODIN & 98.95 & 97.76 & 95.02 & 94.70 \\
        CSI & 92.55 & 95.18 & 90.71 & 91.16 \\
        ARPL & 92.62 & 87.69 & 88.57 & 93.66 \\
        MOS & 74.81  & 73.66 & 70.35 & 94.83 \\
        VOS & 91.56 & 92.18 & 89.68 & 94.31 \\
        LogitNorm & 99.14 & 98.25 & 94.77 & 94.30 \\
        LogitNorm + PixMix & 98.53 & 98.36 & 98.74 & 94.60 \\
        CIDER & 93.30 & 98.06 & 93.71 & / \\
        NPOS & 92.64 & 98.88 & 94.44 & / \\
        \cellcolor{gray!30} \textit{ABN Module V2S} (ours) & \cellcolor{gray!30}93.15  & \cellcolor{gray!30}95.42 & \cellcolor{gray!30}94.88 & \cellcolor{gray!30}97.5 \\
        \hline\hline
        \multicolumn{5}{| c |}{\cellcolor{blue!15}\textbf{Trained + Outliers}}\\ 
        \hline\hline
        OE & 90.22 & \textbf{99.60}  & 97.58 & 94.63 \\
        MCD & 84.22 & 93.76 & 93.35 & 94.95 \\
        UDG & 95.81 & 94.55 & 93.92 & 92.36 \\
        MixOE & 91.66 & 93.82 & 91.84 & 94.55 \\
        \cellcolor{gray!30} \textit{ABN Module V2M} (ours) & \cellcolor{gray!30}\textbf{99.24}  & \cellcolor{gray!30}98.06 & \cellcolor{gray!30}\textbf{99.43} & \cellcolor{gray!30}97.5\\
         \hline\hline
        \end{tabular}
        \caption{AUROC metric for Far-OOD detection on CIFAR10 Benchmark.} 
         \label{benchmark_AUROC_far}
    \end{center}
\end{table}

\begin{table}[t]
    \footnotesize
    \centering
    \begin{center}
        \begin{tabular}{ | l | c | c | c  |}
        \hline
        \textbf{Technique} & \textbf{CIFAR-100} & \textbf{TIN} & \textbf{ID ACC} \\
        \hline\hline
        \multicolumn{4}{| c |}{\cellcolor{blue!15}\textbf{Post-Hoc}}\\ 
        \hline\hline
        OpenMax & 86.91 & 88.32 & 95.06 \\
        MSP & 87.19 & 88.87 & 95.06 \\
        TempScale & 87.17 & 89.00 & 95.06 \\
        ODIN & 82.18 & 83.55 & 95.06 \\
        MDS & 83.59 & 84.81 & 95.06 \\
        MDSEns & 61.29 & 59.57 & 95.06 \\
        RMDS & 88.83 & 90.76 & 95.06 \\
        Gram & 58.33 & 58.98 & 95.06 \\
        EBO & 86.36 & 88.80 & 95.06 \\
        OpenGAN & 52.81 & 54.62 & 95.06 \\
        GradNorm & 54.43 & 55.37 & 95.06 \\
        ReAct & 85.93 & 88.29 & 95.06 \\
        MLS & 86.31 & 88.72 & 95.06 \\
        KLM & 77.89 & 80.49 & 95.06 \\
        VIM & 87.75 & 89.62 & 95.06 \\
        KNN & 89.73 & 91.56 & 95.06 \\
        DICE & 77.01 & 79.67 & 95.06 \\
        RankFeat & 77.98 & 80.94 & 95.06 \\
        ASH & 74.11 & 76.44 & 95.06 \\
        SHE & 80.31 & 82.76 & 95.06 \\
        GEN & 87.21 & 89.20 & 95.06 \\
        \hline\hline
        \multicolumn{4}{| c |}{\cellcolor{blue!15}\textbf{Trained}}\\ 
        \hline\hline
        ConfBranch & 88.91 & 90.77 & 94.88 \\
        RotPred & 91.19 & 94.17 & 95.35 \\
        RotPred + PixMix & \textbf{93.48} & 96.24 & 95.91\\
        G-ODIN & 88.14 & 90.09 & 94.70 \\
        CSI & 88.16 & 90.87 & 91.16 \\
        ARPL & 86.76 & 88.12 & 93.66 \\
        MOS & 70.57 & 72.34 & 94.83 \\
        VOS & 86.57 & 88.84 & 94.31 \\
        LogitNorm & 90.95 & 93.70 & 94.30 \\
        LogitNorm + PixMix & 92.91 & 95.60 & 94.60 \\
        CIDER & 89.47 & 91.94 & / \\
        NPOS & 88.57 & 90.99 & / \\
        \cellcolor{gray!30}\textit{ABN Module V2S} (ours) & \cellcolor{gray!30}69.07 & \cellcolor{gray!30}84.02 & \cellcolor{gray!30}97.5\\
        \hline\hline
        \multicolumn{4}{| c |}{\cellcolor{blue!15}\textbf{Trained + Outliers}}\\ 
        \hline\hline
        OE & 90.54 & \underline{99.11} & 94.63 \\
        MCD & 89.88 & 92.18 & 94.95 \\
        UDG & 88.62 & 91.20 & 92.36 \\
        MixOE & 87.47 & 90.00 & 94.55 \\
        \cellcolor{gray!30}\textit{ABN Module V2M} (ours) & \cellcolor{gray!30}\underline{93.36}  & \cellcolor{gray!30}\textbf{99.66} & \cellcolor{gray!30}97.5\\
         \hline\hline
        \end{tabular}
        \caption{AUROC metric for Near-OOD detection on CIFAR10 Benchmark.
        }
         \label{benchmark_AUROC_near}
    \end{center}
\end{table}

\section{Conclusion}\label{section_conclusion}
In this study, we have addressed the problem of defining a novel approach to robust Out-Of-Distribution detection, studying and defining methods in the context of deepfake detection. Two different approaches that share a common pipeline are proposed. This procedure is composed of two cascade modules. The first is mainly constituted by the In-Distribtion classifier, which in this case addresses the deepfake detection problem, producing the detection outcomes and other auxiliary data that are shared with the second module.
The second module is a custom Abnormality module that makes inferences on which sample belongs to the ID and which to the OOD, employing the classification outputs and the auxiliary data from the first module. The two different approaches differ on the typology of In-Distribution module involved and which kind of auxiliary data is shared with the abnormality module. Specifically are proposed solutions with Convolutional Neural Networks and with Vision Transformers. We have proposed three different Abnormality module network architectures, with different data reduction strategies. Three different modalities of Abnormality module training are covered in the analysis: applying proposed techniques to synthesize OOD data from ID samples, employing real OOD data, and mixing real and synthetic OOD data. We have defined train and test model procedures on a custom version of the CDDB dataset.
Moreover, it’s integrated into the study the CIFAR10 OOD benchmark, to extend the analysis concerning a wider range of available techniques in the field. We have obtained promising performances with both approaches. The Transformer-based strategy, thanks to the Tiny DeiT model, is computationally convenient and provides a faster convergence in the ID classification task. This solution records better OOD detection performances in the Content scenario with respect to the usage of the CNN-based solution. Above all, we point out Abnormality modules V2 and V3, which show remarkable efficacy also for future research. Results comparable to those of the state-of-the-art are obtained involving real OOD data as outliers, which is appreciable in the benchmark results. Despite that, it’s hard to find suitable outlier samples to be included in the training of these OOD detection modules. Indeed, the model performance is strictly related to the quality of these data, and how well they describe distributions different from one learned by the classifier, which could vary based on the task typology. Thus, we believe that the inclusion of only synthetic OOD data to train the models is the most attractive direction for this study since reflects a more general, and easy-use solution. Looking at the content scenario metrics related to real/fake faces, and the Far-OOD benchmark evaluation, we have demonstrated the feasibility of this approach.\\
Possible future research steps could involve new OOD synthetization techniques and novel auxiliary data production for the abnormality module. For example, the inclusion of GAN-generated OOD images, different noise distributions like salt and pepper or speckle noise, images filtered at different frequencies, use of noise fingerprint as new auxiliary data, and so on. Furthermore, This study can be extended to a multi-class classification ID task related to the deepfake techniques, with the possibility of a further extension to a multimodal study or integration of images at higher resolution. Exploring such scenarios might help us comprehend more in-depth how different tasks and configurations affect Out-of-Distribution (OOD) detection performance.

\section*{Acknowledgments}
This study has been partially supported by SERICS (PE00000014) under the MUR National Recovery and Resilience Plan funded by the European Union - NextGenerationEU. The authors are with Sapienza University of Rome, Italy (e-mail: casadei.1952529@studenti.uniroma1.it, maiano@diag.uniroma1.it, amerini@diag.uniroma1.it).

\bibliographystyle{unsrt}  
\bibliography{references}

\end{document}